\documentclass[a4paper,fleqn,twocolumn]{cas-dc}

\usepackage[numbers]{natbib}
\usepackage{graphicx}
\usepackage[linesnumbered,ruled,vlined]{algorithm2e}
\def\tsc#1{\csdef{#1}{\mathrmsc{\lowercase{#1}}\xspace}}
\tsc{WGM}
\tsc{QE}
\tsc{EP}
\tsc{PMS}
\tsc{BEC}
\tsc{DE}

\begin{document}
\let\WriteBookmarks\relax
\def\floatpagepagefraction{1}
\def\mathrmpagefraction{.001}
\shorttitle{Cali-Sketch}
\shortauthors{Xia et~al.}

\title[mode=title]{Cali-Sketch: Stroke Calibration and Completion for High-Quality Face Image Generation from Human-Like Sketches}

\author[1]{Weihao Xia}%
\address[1]{Tsinghua University, China}

\author[2]{Yujiu Yang}[orcid=0000-0002-6427-1024]
\ead{Corresponding author:yang.yujiu@sz.tsinghua.edu.cn}
\address[2]{Tsinghua Shenzhen International Graduate School, Tsinghua University, China}

\author[3]{Jing-Hao Xue}%
\address[3]{Department of Statistical Science, University College London, UK}

\begin{abstract}
Image generation has received increasing attention because of its wide application in security and entertainment. Sketch-based face generation brings more fun and better quality of image generation due to supervised interaction. However, when a sketch poorly aligned with the true face is given as input, existing supervised image-to-image translation methods often cannot generate acceptable photo-realistic face images. To address this problem, in this paper we propose Cali-Sketch, a human-like-sketch to photo-realistic-image generation method. 
Cali-Sketch explicitly models stroke calibration and image generation using two constituent networks: a Stroke Calibration Network (SCN), which calibrates strokes of facial features and enriches facial details while preserving the original intent features; and an Image Synthesis Network (ISN), which translates the calibrated and enriched sketches to photo-realistic face images. 
In this way, we manage to decouple a difficult cross-domain translation problem into two easier steps. Extensive experiments verify that the face photos generated by Cali-Sketch are both photo-realistic and faithful to the input sketches, compared with state-of-the-art methods.
\end{abstract}

\begin{keywords}
Face sketch-to-photo synthesis \sep Image translation \sep Neural network \sep Generative adversarial network\sep
\end{keywords}

\maketitle

\section{Introduction}
\label{intro}

Drawing a sketch is maybe the easiest way for amateurs to describe an object or scene quickly. Compared with photographs or portraits, it does not require technical capture devices or professional painting skills. Generating photo-realistic images from free-hand sketch enables a novice to create images from their imagination, making reality a face or scene otherwise only exist in their dreams. However, the sketches drawn by non-artists are usually simple and imperfect. They are sparse, the lack of necessary details, and strokes do not precisely align with the original images or actual objects. It is hence challenging to synthesize natural and realistic images from such human-like sketches.

\begin{figure*}[pos=t]
\centering
\includegraphics[width=\textwidth]{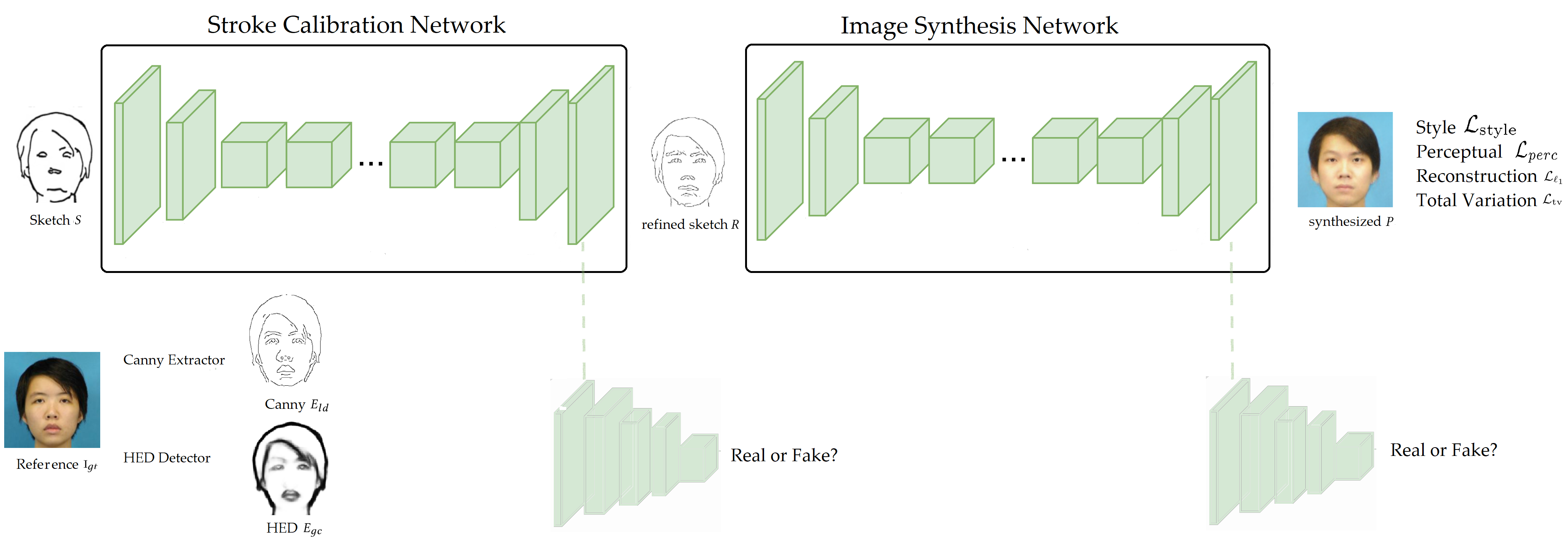}
\caption{The overall architecture of our method. Stroke Calibration Network first calibrates unreasonable strokes and adds necessary details. The modified sketches are then fed into the Image Synthesis Network to produce photo-realistic face images.}
\label{fig:pipeline}
\end{figure*}

Recent progress on image-to-image translation \cite{isola2017image,zhu2017unpaired, Zhou2019BranchGAN,StarGAN2018,wang2018pix2pixHD,Chen2019Quality,XuKW19} has shown that an end-to-end generative adversarial network (GAN) architecture could produce high quality results. A few of them are capable of synthesizing facial photos from sketches, but it requires sketches with precisely and strictly aligned boundaries to produce plausible results. Building such an exquisite large-scale dataset with thousands of image pairs (i.e, face photo and its corresponding sketch drawn by professional portraitists) would be quite time-consuming and expensive. It is much easier to build a dataset of face photos and their corresponding free-hand sketches drawn by amateurs. Technically, given such human-like sketches and photos, the networks proposed for cross-domain translation \cite{isola2017image,zhu2017unpaired,huang2018munit,lee2020drit++} would learn both stroke modification and image generation simultaneously. However, the remarkable stroke and appearance differences between sketches and photos diminish the effectiveness of these networks, thus leading to unpleasant results.

There are some interactive face image modification methods under the framework of image inpainting \cite{jo2019sc,yu2018generative,nazeri2019edgeconnect,liu2019homography}. Given a partially-and-irregularly masked image, they refill the erased regions with strokes provided by the user as guidance. The refilled regions are consistent with input reference strokes and compatible with the whole image. Recent work \cite{jo2019sc} can obtain a realistic synthetic face photo even though the user conducts some modifications and the network tolerates minor error or mismatching. However, to generate an appropriately edited and restored result, a plausible sketch of the original image is still needed by these methods.  When human-like sketches are fed into the model, the results can be unacceptable. Moreover, synthesizing an image from a total sketch is much harder than from a regionally-erased image, since in the latter case the rest edges and colors can significantly help the reconstruction.

Some methods \cite{Chen2018CVPR, lu2018sketch} consider the case of casual free-hand frontal face sketches, where the generated images do not have to strictly align with the input sketches and present more freedom in appearance. But their methods produce blurry and artifactual results. What's more, crucial components and drawing intention of the original sketches such as facial contours and hairstyles are not preserved in the synthesized images.

To address these issues, we propose a novel two-stage generative adversarial network called ``Cali-Sketch``, to realize face photo synthesis from human-like sketches in a unified framework. It explicitly models stroke calibration and image generation using two constituent GANs: a Stroke Calibration Network  ({\it SCN}), which calibrates and completes strokes of facial features and enriches facial details while preserving the original intent features of the painter, and an Image Synthesis Network ({\it ISN}), which transfers the calibrated and completed sketches to face photos. Two GANs are first separately trained for each stage, and then trained jointly.

We focus on image generation from ``human-like sketches", which is less discussed but appears more in real applications. Compared with the aforementioned methods in \cite{isola2017image, jo2019sc,wang2018pix2pixHD, lu2019fcn,liu2018auto}, ours doesn't necessarily desire a sketch well aligned to the original image to generate an appropriately edited and restored result while those methods might produce unacceptable results given such human-like sketches as input.

To preserve facial features and drawing intention, we propose both global contour loss and local detail loss to accomplish necessary stroke modifications and detail improvements. To eliminate artifacts, we also incorporate a perceptual loss and a reconstruction loss in the overall objective function. In this way, we manage to make the final appearance of generated images photo-realistic, while keeping the determinant attributes and drawing intention of the input sketch. Experiments confirm that face images synthesized by our proposed method are natural-looking and visually pleasant without observable artifacts.

To sum up, our key contributions are three-fold:

\begin{itemize}
\item We present the first two-stage human-like face sketch to photo translation. It achieves stroke calibration and image synthesis with two consecutive GANs: SCN and ISN.
\item We propose SCN for necessary stroke calibration and detail completion. To preserve identity and drawing intention during the reconstruction of fine-grained face sketches, we design novel calibration loss functions. Furthermore, when given a free-hand drawn sketch, this network can act as a pre-processing modification module for other tasks using reference sketches such as interactive face image modification.
\item We propose ISN for face sketch-to-image generation. The synthesized face images are both identity-consistent and appearance-realistic.
\end{itemize}

The rest of this paper is organized as follows.
Section \ref{sec:related works} provides an overview of the previous methods and related techniques. Section \ref{sec:method} presents the proposed Cali-Sketch method.
Section \ref{sec:experiments} reports the qualitative and quantitative performance of sketch-based image synthesis experiments using the proposed method, and Section \ref{sec:conclusion} summarizes and concludes the paper.

\section{Related Work}
\label{sec:related works}
\subsection{Photo-Realistic Image Synthesis}
Photo-realistic image synthesis methods have progressed rapidly during the last few years. The goal of image synthesis is to generate photo-realistic and faithful images from sketches or abstract semantic label maps, refer to as label-based image synthesis and sketch-based image synthesis, respectively.

Label-based image synthesis methods \cite{wang2018pix2pixHD, ChenK17Photographic,QiCJK18Synthesis} synthesize image semantically from abstract label maps, such as sparse landmarks or pixel-wise segmentation maps.
\cite{wang2018pix2pixHD} proposes a framework for instance-level image synthesis with conditional GANs. \cite{ChenK17Photographic} proposes a cascade framework to synthesis high-resolution images from pixel-wise labeling maps.

Facial sketch-based image synthesis approaches have been widely developed during the last few years. 
Those existing studies can be broadly classified into two categories: image retrieval based approaches \cite{chen2009sketch2photo,eitz2011photosketcher,chen2013poseshop,song2012image,liang2012face} and deep learning based methods \cite{isola2017image,zhu2017unpaired,Chen2018CVPR,lu2018sketch,wang2018high,gao2017ca-gan}. 
The former mainly has three basic steps: retrieve, select and composite. 
Given a sketch plus overlaid text labels as input, Sketch2Photo \cite{chen2009sketch2photo} and Photo-Sketcher \cite{eitz2011photosketcher} automatically synthesize realistic pictures by seamlessly composing objects and backgrounds based on sketch searching and image compositing. PoseShop \cite{chen2013poseshop} constructs a large segmented character database for human synthesis, where people in pictures are segmented and annotated by actions and appearance attributes. Then human images are composed by feeding given sketches with text labels into the query. Those methods often suffer from heavily blurred effects and tedious inference process.

Deep learning based methods learn the mapping between sketches and photos. The pix2pix \cite{isola2017image} translates precise edge maps to pleasing shoe pictures using conditional GANs. CycleGAN \cite{zhu2017unpaired} proposes cycle-consistent loss to handle the paired training data limitation of pix2pix.
SketchyGAN \cite{Chen2018CVPR} synthesizes plausible images of objects from 50 classes. It aims to synthesis results both photo-realistic and faithful to the intention of given sketches. In this case, the intention was defined as generated images sharing similar poses with input sketches since it is hard to learn human intention.
PhotoSketchMAN \cite{wang2018high} generates face photos iteratively from low resolution to high resolution by multi-adversarial networks. CA-GAN \cite{gao2017ca-gan} proposes to use pixel-wise labelling facial composition information to help face sketch-photo synthesis. Contextual-GAN \cite{lu2018sketch} formulates the task of sketch-image synthesis as the joint image completion. Sketches provide contextual information for completion.

\begin{figure*}[pos=th]
  \centering {\includegraphics[width=\textwidth]{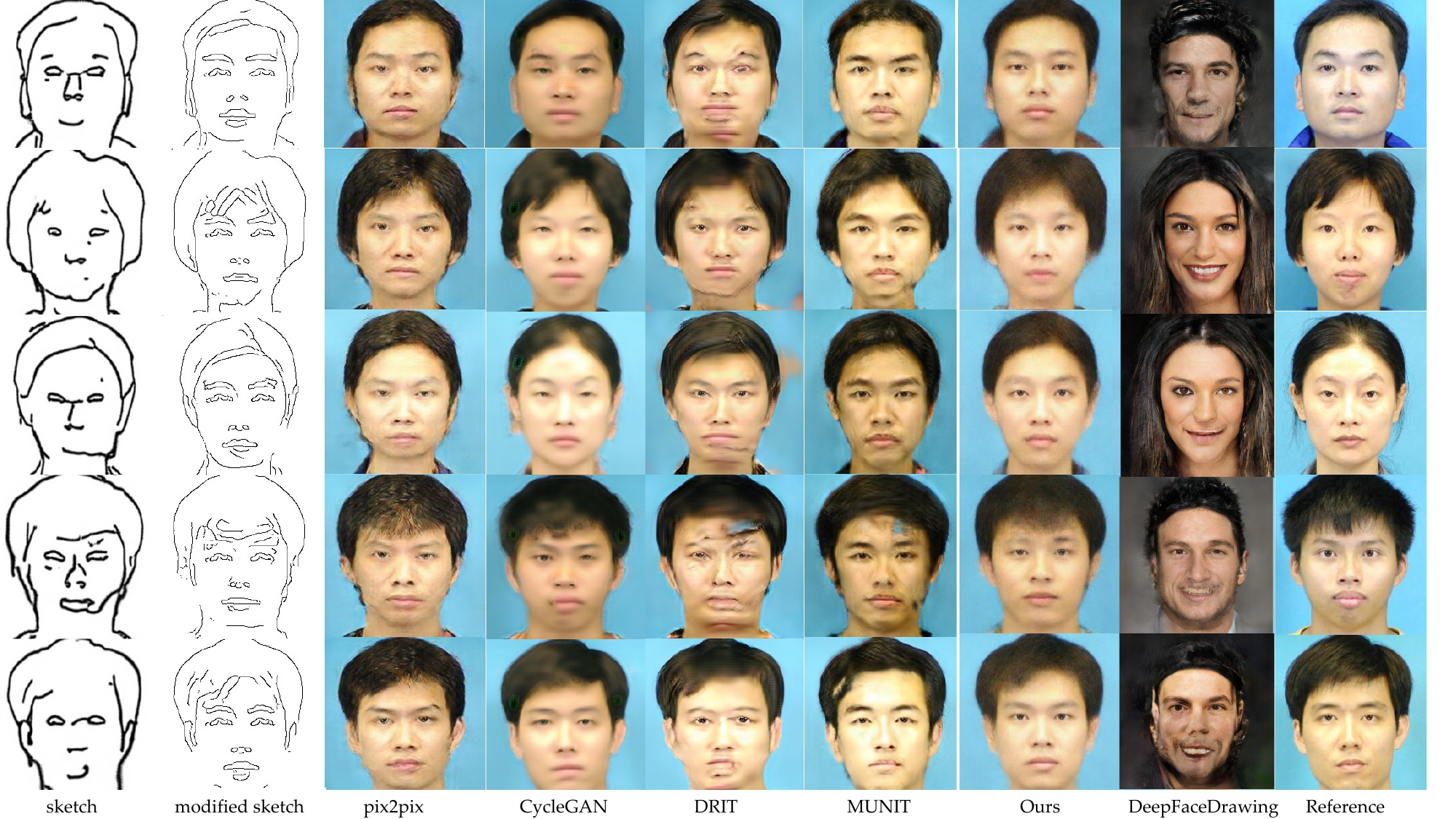}}
  \caption{Qualitative comparison with baselines. We compare our methods with pix2pix \cite{isola2017image}, CycleGAN \cite{zhu2017unpaired}, DRIT \cite{lee2020drit++}, MUNIT \cite{huang2018munit}, \textcolor{blue}{and a very recent sketch-to-image method DeepFaceDrawing~\cite{chen2020DeepFaceDrawing}}. Our approach generates more photo-realistic images. The corresponding image can be recognised easily from a batch of mixed sketches, which means crucial components and drawing intention of original sketches such as facial contours and hair styles are well-preserved in the synthesized images.}
  \label{fig:qualitative comparison with baselines}
\end{figure*}

\subsection{Generative Adversarial Networks}
In recent years, Generative Adversarial Networks (GANs) \cite{Goodfellow2014Generative} have been successfully applied in many computer vision tasks to improve the realism of generated images, such as domain adaption \cite{murez2018image,deng2018image}, super-resolution \cite{wang2018esrgan,xue18Review}. 
They are composed of a generator $G$  and a discriminator $D$. Discriminators try to distinguish the generated fake images, while generators aim to fool discriminators from identifying real images from fake ones. The ideal solution is the Nash equilibrium where $G$ and $D$ couldn't improve their cost unilaterally.

Despite great success, there are still several challenges in GANs including generalization \cite{arora2017generalization,mescheder2017numerics} and training stability \cite{arjovsky2017wasserstein,salimans2016improved}. To alleviate those problems, technologies are proposed to improve GANs. For example, Arjovsky \emph{et al.} \cite{arjovsky2017wasserstein,Gulrajani2017Improved} propose to minimize the Wasserstein distance between model and data distributions. Berthelot \emph{et al.} \cite{berthelot2017began} try to optimize a lower bound of the Wasserstein distances between auto-encoder loss distributions on real and fake data distributions. Mao \emph{et al.} \cite{mao2017least} proposes a least-squares loss for the discriminator, which implicitly minimizes Pearson $\chi^2$ divergence, leading to stable training, high image quality and considerable diversity.

\subsection{Image-to-Image Translation with GANs}
General image-to-image translation methods aim to learn a mapping from the source domain to the target domain. Isola \emph{et al.} \cite{isola2017image} propose a pix2pix framework trained with image pairs and achieve convincing synthetic images on many translation tasks. To handle the limitation of paired images for training,  CycleGAN \cite{zhu2017unpaired}, DualGAN \cite{yi2017dualgan}, DiscoGAN \cite{kim2017learning} present cycle consistency loss to constrain the translation between inputs and translated images.
CSGAN \cite{kancharagunta2019csgan} extends \cite{zhu2017unpaired} with an additive cyclic-synthesized loss between the synthesized image of one domain and the cycled image of another domain.
InstaGAN \cite{mo2018instagan} incorporates instance attribute information for multi-instance transfiguration.
MUNIT \cite{huang2018munit} and DRIT \cite{lee2020drit++} are proposed for one-to-many diverse image translation. ComboGAN \cite{anoosheh2018combogan}  also proposes a multi-component translation method without being constrained to two domains.

\section{Method}
\label{sec:method}
\subsection{Overview}
\begin{figure}[pos=t]
  \centering
  \includegraphics[width=\linewidth]{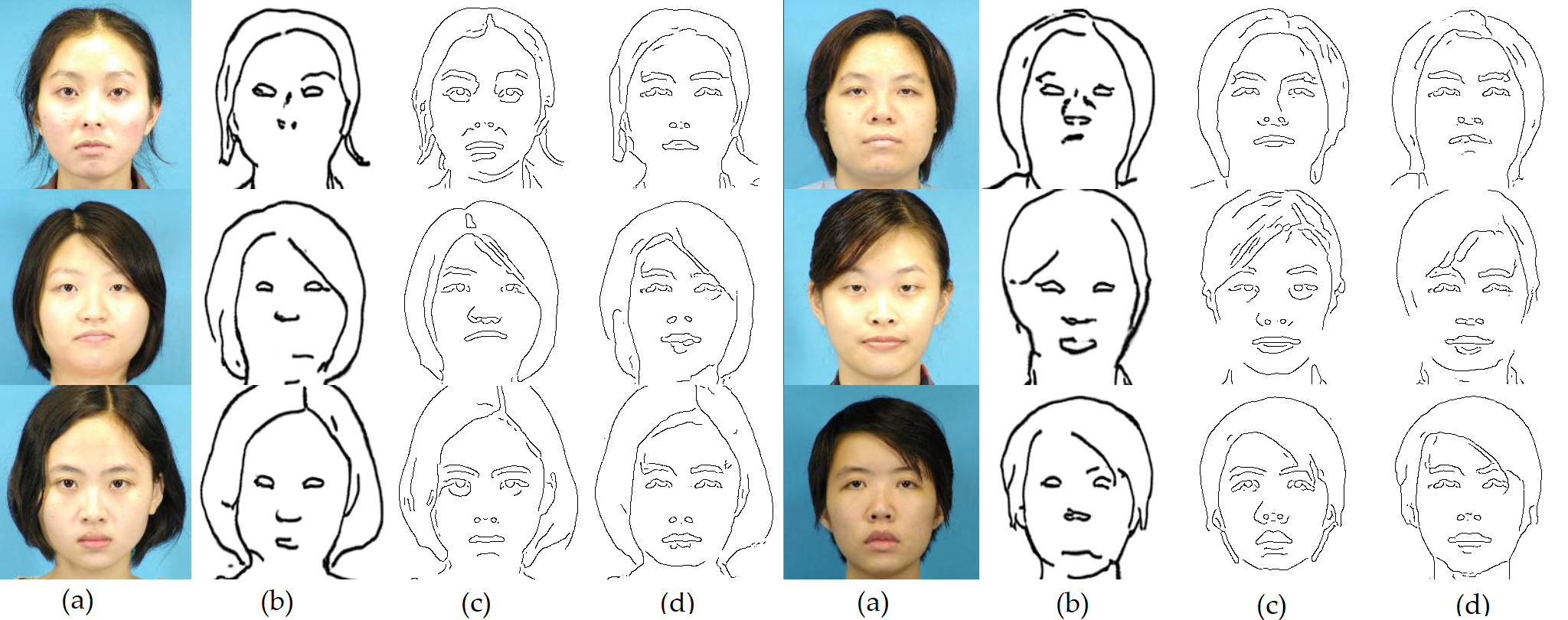}
\caption{Illustration of Stroke Calibration Network (SCN): (a) reference; (b) input sketch; (c) Canny result; (d) modified sketch by our SCN.}
  \label{fig:Illustration of SCN}
\end{figure}

\begin{figure}[pos=t]
\centering
\includegraphics[width=\linewidth]{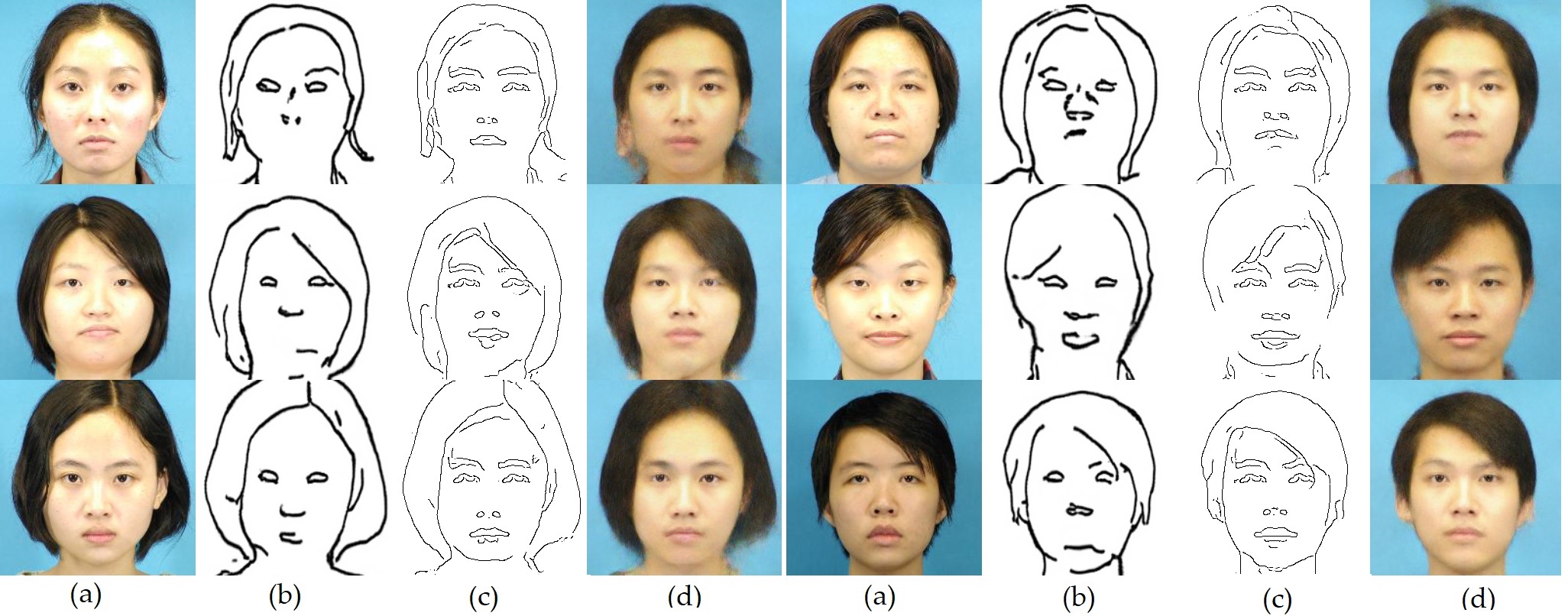}
\caption{Illustration of Image Synthesis Network (ISN): (a) reference; (b) original input sketch; (c) input sketch modified by our SCN. (d) generated image by our ISN.}
\label{fig:Illustration of ISN}
\end{figure}

Our goal is to realize face photo synthesis from a human-like sketch.  Consider two data collections from different domains, $\textbf{S}\subset {{I}^{H\times W\times1}} $ referring to input sketch domain and  $\textbf{P}\subset {{I}^{H\times W\times3}}$ referring to output photo domain. $I^{H\times W\times N}$ represents an image of height $H$, width $W$ and channel $N$. Converting a face sketch from source domain $\textbf{S}$ to an image in the target photo domain $\textbf{P}$ can be referred to as ${G: \textbf{S} \to \textbf{P}}$ . This is a typical cross-domain image translation problem but we could not directly learn the mapping by existing image-to-image translation methods. Instead, we decompose this translation into two stages: 1) Stroke Calibration Network named {\it SCN}, and  2) Image Synthesis Network named {\it ISN}. Let $G_1$ and $D_1$ be the generator and discriminator of SCN, $G_2$ and $D_2$ be the generator and discriminator of ISN, respectively. As shown in Figure \ref{fig:pipeline}, the input sketch $\textbf{S}$ is first put into SCN to get the refined sketch $\textbf{R}$ after stroke calibration and detail completion, which is then fed into ISN to generate a photo-realistic face image $\textbf{P}$. We first train Stroke Calibration Network and Image Synthesis Network separately until the losses plateau, and then train them jointly in an end-to-end way until convergence. Qualitative comparison with baselines is demonstrated in Figure \ref{fig:qualitative comparison with baselines}. Illustrations of SCN and ISN are shown in Figure \ref{fig:Illustration of SCN} and \ref{fig:Illustration of ISN}, respectively. Training details and network architecture can be found in Section \ref{sec:experiment settings}.

\subsection{Stroke Calibration Network}
\label{sec:SCN}
Stroke Calibration Network aims to modify inconsequent strokes and enrich necessary details of input sketch. Let $\mathbf{S}$ be input sketches. Ground truth face photos and their edge counterparts will be denoted as  $\mathbf{I}_{gt}$ and $\mathbf{E}_{gt}$. 
The mapping from human-like sketches  $\mathbf{S}$ to the modified ones $\mathbf{R}$  can be denoted as ${G_{1}: \mathbf{S} \to \mathbf{R}}$:
\begin{equation}
 \mathbf{R}=G_{1}\left(\mathbf{S}, \mathbf{E}_{gt}\right)
\end{equation}
where $\mathbf{E}_{gt}$ are composed of two components: global contours $\mathbf{E}_{gc}$ and local details $\mathbf{E}_{ld}$.

To modify inconsequent strokes and enrich necessary details, we introduce a novel calibration loss $\mathcal{L}_{CL}$ which consists of global contour loss and local detail loss. Global contour loss aims to modify inconsequent strokes and local detail loss enriches necessary details.

We define both losses based on the feature matching loss \cite{wang2018pix2pixHD}. 
Feature representations of real and synthesized images extracted from multiple layers of discriminator are then used to calculate the feature matching loss as
\begin{equation}
\mathcal{L}_{CL}=\mathbb{E}\left[\sum_{i=1}^{T} \frac{1}{N_{i}}\left\|D_{1}^{(i)}\left(\mathbf{E}_{gt}[j]\right)-D_{1}^{(i)}\left(\mathbf{R}\right)\right\|_{1}\right],
\end{equation}
where $T$ is the index of the final convolution layer of the discriminator, $N_i$ is the number of elements in the $i$-th activation layer, $\mathbf{E}_{gt}[j], j\in\{ 0,1\}$ represents global contour or local detail, and  $D^{(i)}_1$ is the activation in the $i$-th layer of the discriminator. 
In our experiments, global contour and local detail are implemented by HED \cite{xie15hed} and Canny \cite{canny1987computational} edge map, respectively. This calibration loss can stabilize training by forcing the generator to produce natural statistics at different scales \cite{wang2018pix2pixHD}.

For stable training, high image quality and considerable diversity as discussed in Section \ref{sec:related works}, we use the least-squares GAN \cite{mao2017least} in our experiment. Thus, $\mathcal{L}_{\mathrm{adv}, \mathrm{SCN}}$ can be formulated as
\begin{equation}
\begin{aligned}
\min _{D_{1}} \mathcal{L}_{\mathrm{adv}, \mathrm{SCN}}\left(D_{1}\right)=&\frac{1}{2} \mathbb{E}_{\boldsymbol{x} \sim p{\mathrm(\boldsymbol{x})}}\left[\left(D_{1}(\boldsymbol{x})-b\right)^{2}\right] +\\
& \frac{1}{2} \mathbb{E}_{\boldsymbol{z} \sim p_{\boldsymbol{z}}(\boldsymbol{z})}\left[\left(D_{1}\left(G_{1}(\boldsymbol{z})\right)-a\right)^{2}\right] \\
\min _{G_{1}} \mathcal{L}_{\mathrm{adv}, \mathrm{SCN}}\left(G_{1}\right)=  &\frac{1}{2} \mathbb{E}_{\boldsymbol{z} \sim p_{z}(\boldsymbol{z})}\left[\left(D_{1}\left(G_{1}(\boldsymbol{z})\right)-c\right)^{2}\right],
\end{aligned}
\end{equation}
where a, b, c denote the labels for fake data and real data and the value that G wants D to believe for fake data, respectively. In our experiment, $x$ are ground truth images and $z$ are input sketches sampled from distribution $p{\mathrm(\boldsymbol{z})}$.

The total loss of Stroke Calibration Network combines an improved adversarial loss and calibration loss as
\begin{equation}
\min \limits_{G_{1}} \max \limits_{D_{1}} \mathcal{L}_{G_{1}}=\min \limits_{G_{1}}\left(\max \limits_{D_{1}}\left(\mathcal{L}_{\mathrm{adv}, \mathrm{SCN}}\right)+\lambda \mathcal{L}_{CL}\right),
\end{equation}
where $\lambda$ is regularization parameters controlling the importance of two terms. We set $\lambda=10$ in the experiments.

\subsection{Image Synthesis Network}
\label{sec:ISN}
After stroke calibration and detail completion, the refined sketch $\textbf{R}$ is then fed into Image Synthesis Network to generate photo-realistic face photo $P$. This translation process from the refined sketch $\textbf{R}$ to the photo-realistic face image $\textbf{P}$ can be defined as  ${G_{2}: \textbf{R} \to \textbf{P}}$.
The output image should yield both high sketch identification similarity and favourable perceptual quality, while sharing the same resolution with the input sketch:

\begin{equation}
\mathbf{P}=G_{2}\left(\mathbf{R},\mathbf{I}_{gt}\right)
\end{equation}

We train this image synthesis network with a joint loss, which consists of five terms: an $\ell_{1}$ reconstruction loss $\mathcal{L}_{\ell_{1}}$, adversarial loss $\mathcal{L}_{adv, 2}$ , perceptual loss $\mathcal{L}_{\mathrm{percep}}$, style loss $\mathcal{L}_{\mathrm{style}}$ and total variation loss $\mathcal{L}_{\mathrm{tv}}$:
\begin{equation}
\mathcal{L}_{G_{2}}=\lambda_{1} \mathcal{L}_{\ell_{1}}+\lambda_{2} \mathcal{L}_{\mathrm{adv}, \mathrm{ISN}}+\lambda_{3} \mathcal{L}_{\mathrm{percep}}+\lambda_{4} \mathcal{L}_{\mathrm{style}}+\mathcal{L}_{\mathrm{tv}}
\end{equation}

Reconstruction loss $\mathcal{L}_{\ell_{1}}$ minimizes the differences between reference and generated images:
\begin{equation}
\mathcal{L}_{\ell_{1}}=\mathbb{E}\big[\sum_{i} \frac{1}{N_{i}}\big\|\mathbf{I}_{gt}-\mathbf{P}\big\|_{1}\big]
\end{equation}

Perceptual loss $\mathcal{L}_{\mathrm{percep}}$ is proposed by Johnson \emph{et al.} \cite{Johnson2016Perceptual} based on perceptual similarity. It is originally defined as the distance between two activated features of a pre-trained deep neural network. Here we adopt a more effective perceptual loss which uses features before activation layers \cite{wang2018esrgan}. These features are more dense and thus provide relatively stronger supervision, leading to better performance:
\begin{equation}
\mathcal{L}_{\mathrm {percep}}=\mathbb{E}[\sum_{i} \frac{1}{N_{i}}\|\phi_{i}(\mathbf{I}_{gt})-\phi_{i}(\mathbf{P})\|_{1}],
\end{equation}
where $\phi_{i}$ donates the feature maps before activation of the VGG-19 network pre-trained  for image classification.

Style loss $\mathcal{L}_{\mathrm{style}}$ is adopted in the same form as in the original work~\cite{Johnson2016Perceptual}, which aims to measure differences between covariance of activation features:
\begin{equation}
\mathcal{L}_{\mathrm {style}}=\mathbb{E}_{j}[\|G_{j}^{\phi}(\mathbf{I}_{gt})-G_{j}^{\phi}(\mathbf{P})\|_{1}],
\end{equation}
where $G_{j}^{\phi}$  represents the Gram matrix constructed from feature maps $\phi_{j}$.

Total variation loss is based on the principle that images with unrestrained and possibly spurious detail have high total variation. According to this, reducing the total variation of an image subject to it being a close match to the original image, removes unwanted noises while enforcing spatial smoothness and preserving important details such as edges. It is defined on the basis of the absolute gradient of generated images:
\begin{equation}
\centering
\mathcal{L}_{\mathrm {tv}}=\|{\nabla_x \mathbf{P}}-{\nabla_y \mathbf{P}}\|_{1}.
\end{equation}
For experiments, we use $\lambda_{1}=1$, $\lambda_{2}=0.1$, $\lambda_{3}=0.1$, and $\lambda_{4}= 200$.

\section{Experiments}
\label{sec:experiments}
\subsection{Training Data}
Appropriate and adequate training data is important for network performance. Since it is infeasible to collect large-scale paired images and sketches, most existing free-hand sketch based image synthesis methods generate sketches automatically from images.

To exhibit different styles of free-hand sketches and to improve the network generality, we augment training data by adopting multiple different styles of input sketches. Specifically, we generate four different free-hand sketch styles in total. We use the XDoG edge detector \cite{WinnemollerKO12} and Photocopy effect in Photoshop to generate two styles. To better resemble hand-drawn sketches, we simplified the edge images using \cite{edgar2016sketch} as in \cite{lu2018sketch}. We also use photo-sketch \cite{LIPS2019} to generate the desired face sketches. This recent method generates imperfect alignment contour sketches of input images.
The human-like sketches should be sparse and contain these wrong edges. That's why the Canny algorithm \cite{canny1987computational} shouldn't be chosen to get input sketches. Those edges generated by Canny are solid and well-aligned with input images.
To show the effectiveness and efficiency of our approach, the CUHK Face Sketch Database \cite{wang2009face} is used in our experiment for its appropriateness and popularity. We use its $256\times256\times3$ resized and cropped version.

\begin{figure}[pos=th]
  \centering
  \includegraphics[width=\linewidth]{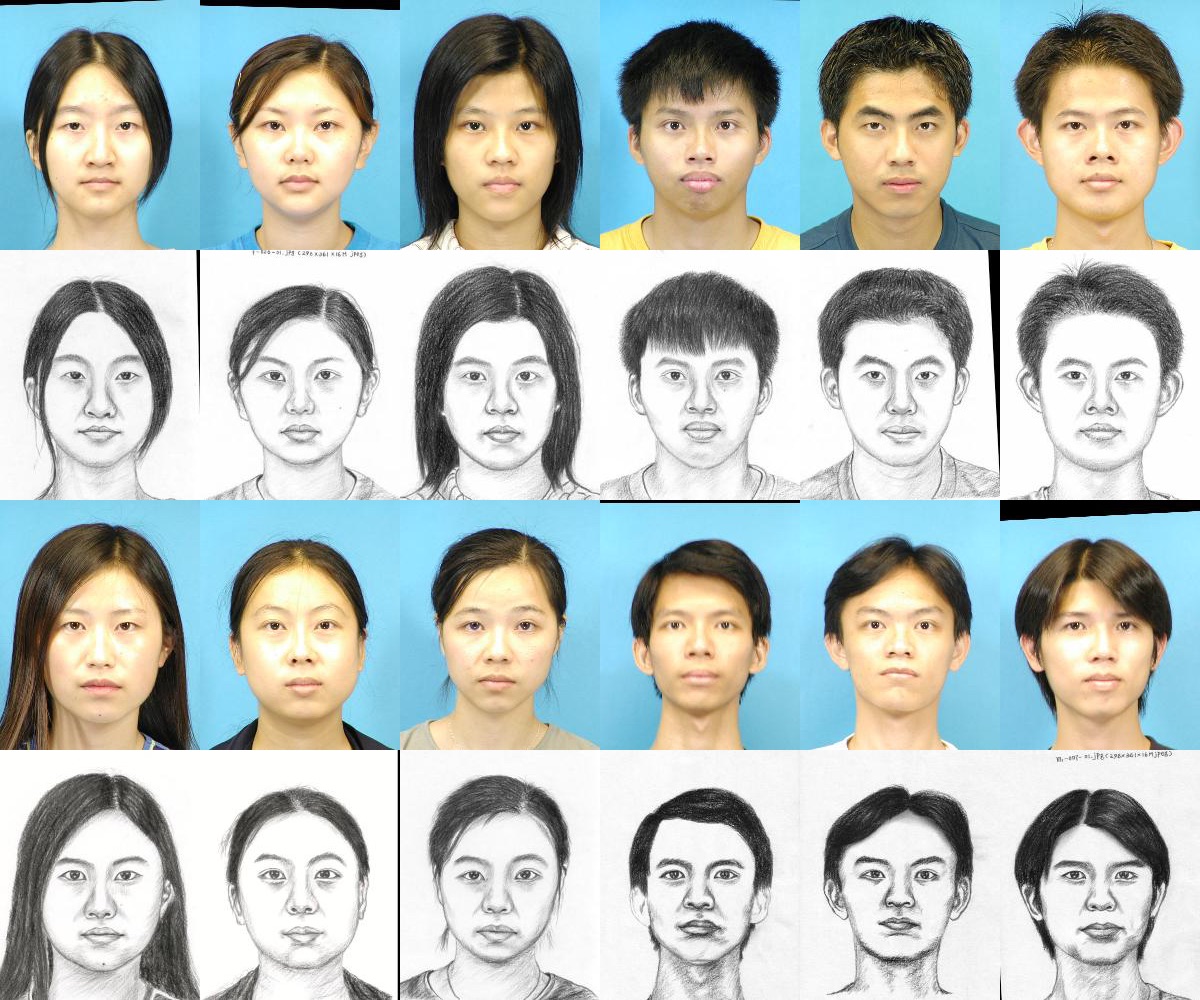}
  \caption{Illustration of well-drawn sketches from \cite{wang2009face}. Best viewed in color.}
  \label{fig:well-drawn-sketch}
\end{figure}

\begin{figure}[pos=t]
  \centering
  \includegraphics[width=\linewidth]{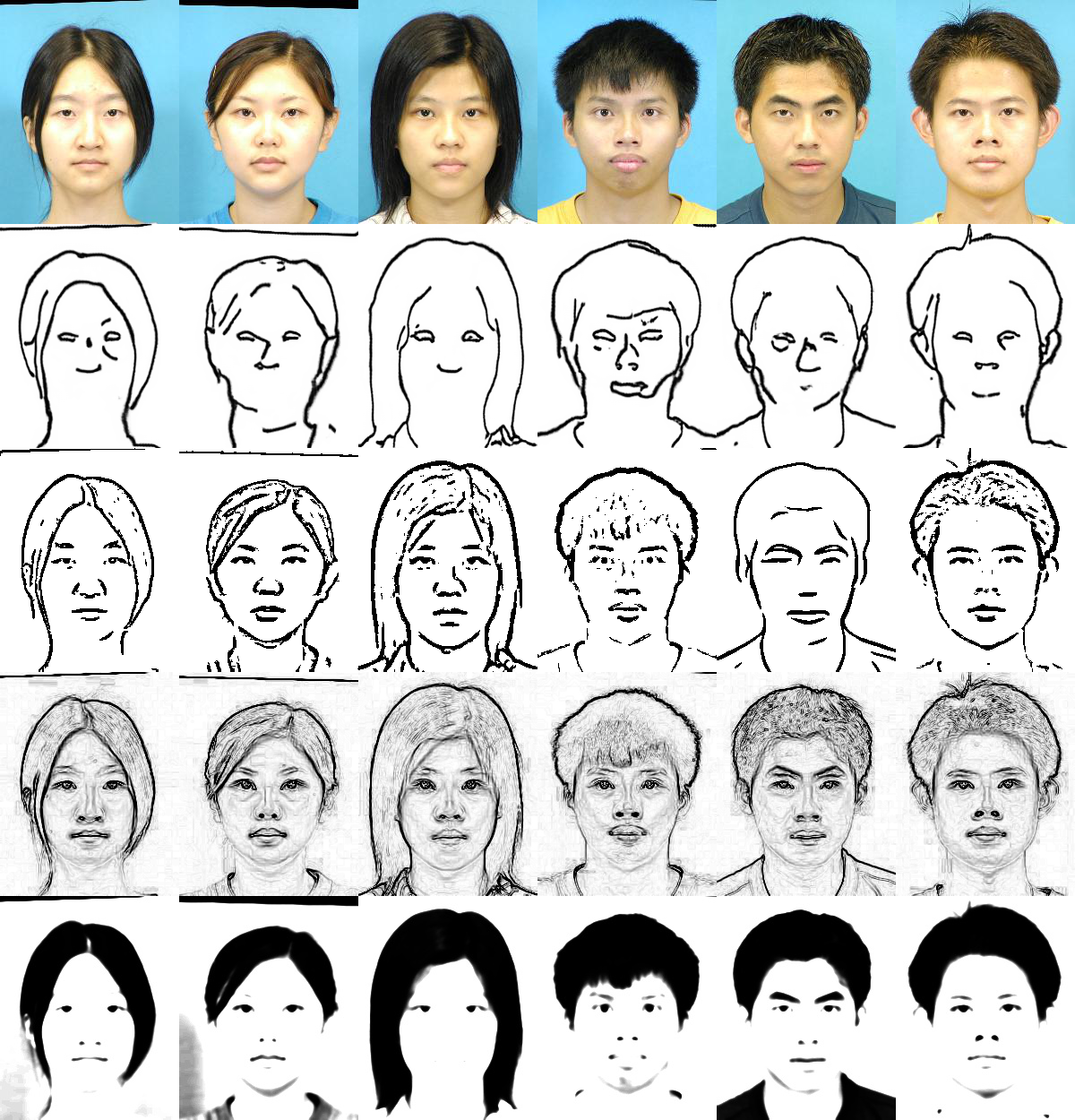}
  \caption{Illustration of four different free-hand sketch styles: photo-sketch \cite{LIPS2019}, XDoG \cite{edgar2016sketch}, Photocopy of Photoshop and FDoG \cite{WinnemollerKO12}.}
  \label{fig:sketch_style}
\end{figure}

Figure \ref{fig:well-drawn-sketch} are illustration of well-drawn sketches from \cite{wang2009face}. These sketches are drawn by the artist. Compared with human-like sketches in Figure \ref{fig:sketch_style}, the well-depicted sketches capture the most distinctive characteristics of human faces and are faithful to the original face images. We often can easily recognize a person from the corresponding sketch. The free-hand sketches are often sparse, deformed, the lack of necessary strikes or details and lines do not precisely align to the real face images, sometimes even the lack of necessary lines in the area of mouth or jaw as illustrated in Figure \ref{fig:sketch_style}.

The ground truth sketches for Stroke Calibration Network are generated using  Canny algorithm \cite{canny1987computational} and Holistically-nested Edge Detection (HED) edge detector \cite{xie15hed}. Specifically, we extract HED from images after histogram equalization to avoid the interference of light. Thus, we generate a desired new dataset consisted of high-quality face photos and corresponding human-like face sketches, Canny together with HED edges.

\subsection{Experiment Settings}
\label{sec:experiment settings}
\textbf{Network architecture.}
Inspired by recent image translation studies, our generators follow an encoder-decoder architecture similar to the method proposed by Johnson \emph{et al.} \cite{Johnson2016Perceptual}. 
Each of the generators consists of two down-sampling encode layers, followed by eight residual blocks\cite{He2016Deep} and two up-sampling decoders. Skip connections are added to concatenate previous layers with the same spatial resolution. We replace regular convolutions in the residual blocks with dilated convolutions with dilation factor two to obtain large receptive fields. 
Our discriminators are based on the SN-PatchGAN~\cite{yu2018free} architecture, which determines whether or not overlapping image patches of a certain size are real. 
Spectral normalization~\cite{miyato2018spectral} is introduced for rapid and stable training and helps produce high-quality results.

Notice that here we did not deliberately design the structure of the Image Synthesis Network (SCN). 
In fact, we adopt a quite simple structure of SCN to show that it is easy to generate satisfactory results from the calibrated sketches even the sktech2image synthesis network is not deliberately designed. For more details about the indispensability of Stroke Calibration Network and scalability of Image Synthesis Network, refer to Section \ref{sec:impact of SCN} and \ref{sec:scalability of ISN} respectively.

\textbf{Training strategy.}
The training strategy is demonstrated in Algorithm \ref{alg:Training strategy}. Forward and backward in Algorithm \ref{alg:Training strategy} represent forward propagation and back propagation respectively. The forward process includes steps of passing the input through the network layers and calculating the actual output and losses of the model. The backward process back-propagates errors and updates the weights of the network. We refer corresponding operations to as forward and backward for simplicity and emphasize that our method is an end-to-end method with three-stage training. $N_1, N_2, N_3$ are iteration numbers which are large enough to guarantee convergence.
Firstly, we train our Stroke Calibration Network $G_1$ using the Canny and HED edges as supervision with a $10^{-4}$ learning rate. Meanwhile, we train Image Synthesis Network $G_2$ using Canny $\odot$ HED as input refined sketches and ground truth face images as supervision with the same  $10^{-4}$ learning rate. Here, $\odot$ denotes the Hadamard product. We then decrease the learning rate to $10^{-5}$ and jointly train both $G_1$ and $G_2$ in an end-to-end way until convergence.  Discriminators are trained with a learning rate of one-tenth of the generators' according to different training stages. Both networks are trained with resized $256 \times 256$ images with a batch of $8$.

\begin{algorithm}[t]
\DontPrintSemicolon

\textbf{Stage 1: SCN training}

\KwIn{$\mathbf{S}$, free-hand sketch. }
\KwOut{$\mathbf{R}$, Refined sketch }

\While{$n \leq N_1$ }{
$\mathbf{R}$, $\mathcal{L}_{G_{1}}$, $\mathcal{L}_{D_{1}}$ =  $G_1$.forward$ (\left(\mathbf{S}, {\mathbf{E}}_{gt}\right))$\;
$G_1$.backward
}

\textbf{Stage 2: ISN training}

\KwIn{$\mathbf{R}$, Canny$\odot$ HED.}
\KwOut{$\mathbf{P}$, Generated face image. }

\While{$n \leq N_2$}{
 $\mathbf{P}$, $\mathcal{L}_{G_{2}}$, $\mathcal{L}_{D_{2}}$ =  $G_2$.forward$ (\left(\mathbf{R}, {\mathbf{I}}_{gt}\right))$\;
$G_2$.backward
}
\textbf{Stage 3: Joint training}

\KwIn{$\mathbf{S}$, free-hand sketch.}
\KwOut{$\mathbf{P}$, Generated face image. }

\While{$n \leq N_3$}{
 $\mathbf{R}$, $\mathcal{L}_{G_{1}}$, $\mathcal{L}_{D_{1}}$ =  $G_1$.forward$ (\left(\mathbf{S}, {\mathbf{E}}_{gt}\right))$\;
 $\mathbf{P}$, $\mathcal{L}_{G_{2}}$, $\mathcal{L}_{D_{2}}$.forward$ (\left(\mathbf{R}, {\mathbf{I}}_{gt}\right))$\;
$G_1$.backward

$G_2$.backward
}
\caption{{\sc Training strategy}}
\label{alg:Training strategy}
\end{algorithm}

\textbf{Evaluation metrics.} For our task of face image synthesis from human-like sketches, we use two kinds of evaluation metrics: similarity metrics and perceptual scores.
We apply the widely used full reference image quality assessment metrics such as PSNR, SSIM as similarity metrics. 
Given two images $I, I^{\prime}\subset I^{H \times W \times C}$, the peak signal-to-noise ratio (PSNR) are defined as
\begin{equation}
\mathrm{PSNR}=10 \log _{10}{\left(\frac{\mathrm{L}^{2}}{\mathrm{MSE}}\right)},
\end{equation}
where $\mathrm{L}$ is usually 255, $\|\cdot\|_{F}^{2}$ is the Frobenius norm and $\mathrm{MSE}=\frac{1}{HWC}\|I-I^{\prime}\|_{F}^{2}$.  The structural similarity index (SSIM) is defined as
\begin{equation}
\mathrm{SSIM}(I, I^{\prime})=\frac{2 \mu_{I} \mu_{I^{\prime}}+k_{1}}{\mu_{I}^{2}+\mu_{I^{\prime}}^{2}+k_{1}} \cdot \frac{\sigma_{I I^{\prime}}+k_{2}}{\sigma_{I}^{2}+\sigma_{I^{\prime}}^{2}+k_{2}},
\end{equation}
where $\mu_{I}$ and $\sigma_{I}^{2}$ is the mean and variance of $I, \sigma_{I \hat{I}}$ is the
covariance between $I$ and $\hat{I},$ and $k_{1}$ and $k_{2}$ are constant
relaxation terms. A highest score indicates a more structurally similar face for a given sketch.

For perceptual scores, we use Fr$\acute{e}$chet Inception distance (FID) \cite{heusel2017gans}. The FID is defined using the Fr$\acute{e}$chet distance between two multivariate Gaussians:
\begin{equation}
\mathrm{FID}=\left\|\mu_{r}-\mu_{g}\right\|^{2}+\operatorname{Tr}\left(\Sigma_{r}+\Sigma_{g}-2\left(\Sigma_{r} \Sigma_{g}\right)^{1 / 2}\right),
\end{equation}
where $X_{r} \sim \mathcal{N}\left(\mu_{r}, \Sigma_{r}\right)$ and $X_{g} \sim \mathcal{N}\left(\mu_{g}, \Sigma_{g}\right)$ are the 2048-dimensional activations of the Inception-v3 pool3 layer for real and generated samples respectively
The lowest FID means it  achieves the most perceptual results.

\subsection{Baselines}
We perform the evaluation on the following baseline methods:

\textbf{Pix2pix} \cite{isola2017image} is an early work of image-to-image translation. It achieves good photo results on edge-to-photo generation, and the models trained on automatically detected edges can generalize to human drawings.

\textbf{CycleGAN} \cite{zhu2017unpaired} achieves unsupervised image-to-image translation via cycle-consistent loss.

\textbf{DRIT} \cite{lee2020drit++} is a recent work, which realizes diverse image-to-image translation via disentangled content and attribute representations of different domains. Experiment on the edge-to-shoes dataset shows it can produce both realistic and diverse images.

\textbf{MUNIT}~\cite{huang2018munit} is the state-of-the-art unsupervised multi-domain image-to-image translation framework. It achieves quality and diversity comparable to the state-of-the-art supervised algorithms on the task of edge-to-shoes/handbags.

\textbf{DeepfaceDrawing}~\cite{chen2020DeepFaceDrawing}~\textcolor{blue}{is a recent state-of-the-art sketch-to-image framework. The key idea of their method is to implicitly model the shape space of plausible face images and synthesize a face image in this space to approximate an input sketch in a local-to-global way.}

\textcolor{blue}{For fair evaluation, all baselines are retrained with the sketch-image pairs except for DeepFaceDrawing~\cite{chen2020DeepFaceDrawing}, which we use their online demo~\footnote{http://geometrylearning.com/DeepFaceDrawing/} since the training scripts are not available.
}
We do not compare with the recent human-drawn sketch to image method \cite{yang2020sketch} since its implementation is not publicly available.

\subsection{Comparison Against Baselines}
\label{sec:comparison against baselines}
\textbf{Qualitative evaluation.} Qualitative comparison with baselines are demonstrated in Figure~\ref{fig:qualitative comparison with baselines}. The results produced by pix2pix~\cite{isola2017image} all have obvious artifacts. All facial features suffer from shape distortion to a degree, especially the facial and ear contours on the first and fifth rows. CycleGAN~\cite{zhu2017unpaired} produces the most similar face with the reference, but its results are blurry and unpleasing. There are two or more visible spots in the area of hair. The contours of face images generated by DRIT~\cite{lee2020drit++} are aligned with their lines of the input sketches, which notably deteriorate the image quality. MUNIT~\cite{huang2018munit} could produce relatively visually realistic and qualitatively consistent results. however, they are more like oil paintings rather than photos.

Compared with baseline methods, our approach generates high-quality images. The generated human face images are more photo-realistic. The corresponding image can be recognized easily from a batch of mixed sketches, which means crucial components and drawing intention of original sketches like facial contours, hairstyles are well-preserved in the synthesized images.

\textbf{Quantitative comparison.} Quantitative evaluation with baselines is shown in Table~\ref{Quantitative comparison}. For PSNR and SSIM, CycleGAN~\cite{zhu2017unpaired} achieves the highest structural similarity, and our method ranks the second. For the task of sketch-to-image generation, the similarity is not that important, since there are no corresponding real face images as reference for most free-hand drawn sketches. What really matters is whether generated images are photo-realistic or not.

Fr$\acute{e}$chet Inception Distance (FID) is calculated by computing the Fr$\acute{e}$chet distance between two Gaussians of feature representations extracted from the pre-trained inception network \cite{szegedy2016rethinking}. It is not only a measure of similarity between two datasets of images, but also shown to correlate well with the human visual judgement of image quality. Due to the above advantages, FID \cite{heusel2017gans} is most often used to evaluate the quality of images generated by Generative Adversarial Networks. As shown in Table \ref{Quantitative comparison}, our method achieves the lowest FID score, which means that our method produces the best results in both perceptual judgement and high-level similarity.

\begin{table}
\centering
\caption{\textcolor{blue}{Perfomance as PSNR, SSIM and FID on the CUHK dataset. The \textbf{best} and \underline{second} best results are highlighted in each column. For details refer to Section \ref{sec:comparison against baselines}}.}
\label{Quantitative comparison}
\begin{tabular}{cccc}
\toprule
Method &PSNR &SSIM &FID\\
\midrule
pix2pix~\cite{isola2017image} &18.83 &0.7554 &\underline{76.90} \\
CycleGAN~\cite{zhu2017unpaired} &\textbf{24.21} &\textbf{0.8508} &80.17 \\
MUNIT~\cite{huang2018munit} & 17.23 &0.7515 &78.57 \\
DRIT~\cite{lee2020drit++} & 16.14 &0.7047 &109.83 \\
\textcolor{blue}{DeepFaceDrawing~\cite{chen2020DeepFaceDrawing}} &13.21 &0.3751 &97.36\\
Ours &\underline{20.25}  &\underline{0.8006} &\textbf{58.43} \\
\bottomrule
\end{tabular}
\end{table}

\subsection{Ablation Studies}
\subsubsection{The choice of contour and detail}
There are many choices of contour and detail for our methods such as edges, boundaries and contours. These are a few differences between them. Edge maps are precisely aligned to object boundaries, and they usually contain more information about details and backgrounds. Boundaries pay more attention to external lines. Contours contain object boundaries, salient internal and background edges. Contours can be obtained by the boundary contour edge extractors like HED \cite{xie15hed}, COB \cite{maninis2018cob}, RCF \cite{liu2017richer,liu2019richer}, or similar to pix2pixHD \cite{wang2018pix2pixHD}, simplified from the face parsing semantic labels. For a face image, contours are more like facial feature boundaries.

Since sketches are the approximate outline of the objects with spatial transformations and deformed strokes~\cite{jing2018stroke}, we need to modify its strokes and add more details before synthesis. Contours and edges are respectively responsible for stroke calibration and detail completion. 
We will illustrate the reasons in the next part.
In our experiment, we choose HED as global contour and Canny as ground truth local detail for simplicity.

\subsubsection{The impact of Stroke Calibration Network}
\label{sec:impact of SCN}
\begin{figure}[pos=h]
  \centering
  \includegraphics[width=\linewidth]{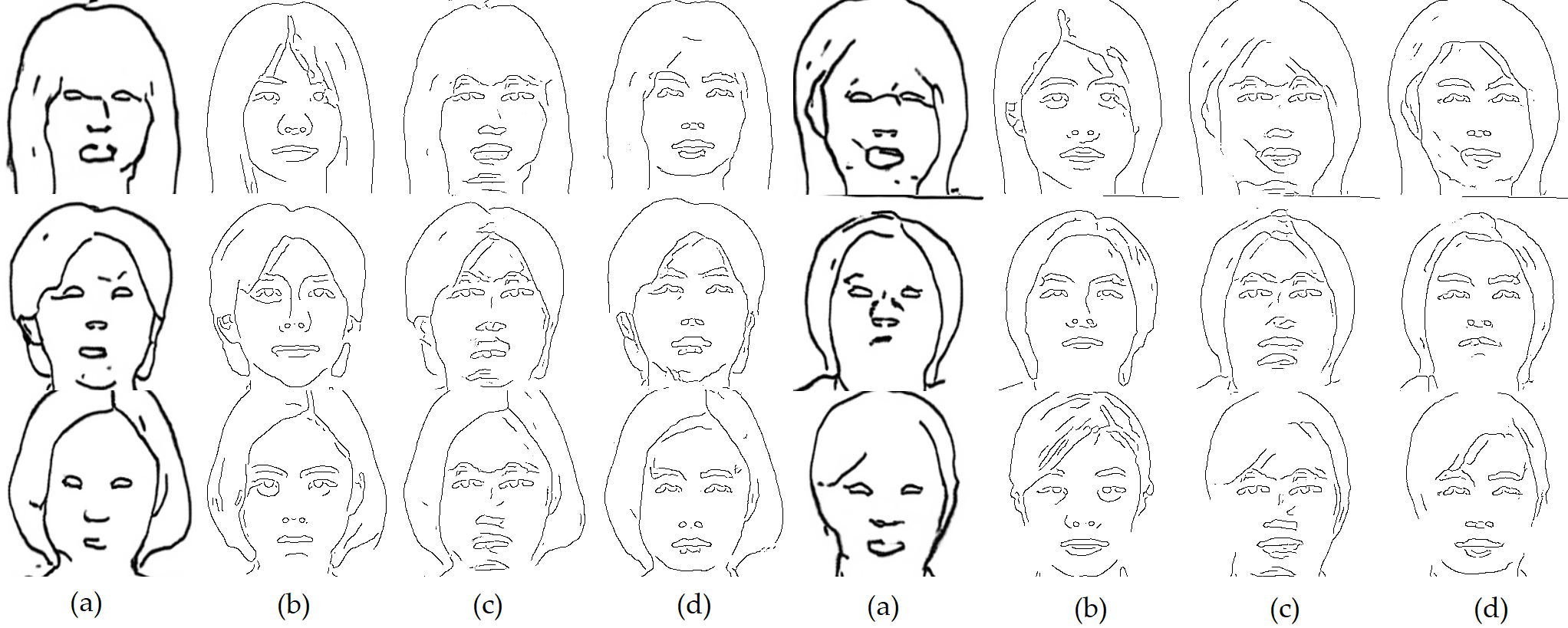}
  \caption{The impact of Stroke Calibration Network and with or without global contour. (a) input sketch. (b) canny edge. (c) result with only local detail loss. (d) result with both local detail loss and global contour loss.}
  \label{fig:impact_scn}
\end{figure}

We have tested directly applying the pix2pix to generate face images from human-like sketches, but found the training unstable and the quality of results unsatisfactory. The original sketches are deformed and sometimes lack of necessary lines in the area of mouth or jaw,  as shown in Figure \ref{fig:Illustration of SCN} and Figure \ref{fig:impact_scn}. It inspires us to modify strokes and add essential details before image synthesis. Edges like the Canny detector can act as ground truth for the training of this process. The refined sketches are more visually favourable and consistent with the original identity, as shown in the third column of Figure \ref{fig:impact_scn}.

However, it also demonstrates that only being supervised by Canny is not enough. It results in unwanted strokes in the area of the eyebrow mouth or jaw, and even changes the shape of the eyes. Stroke calibration should be superior to detail enrichment. We want the Stroke Calibration Network to modify strokes without changing holistic properties like facial contours and hairstyles. So we add both the contours as the global constraint and the edges as a local constraint. As shown in the fourth column of Figure \ref{fig:impact_scn}, it calibrates unreasonable strokes and preserves original properties.

Table \ref{tab:impact of SCN} shows the accuracy of our Stroke Calibration Network. We measure precision and recall with Canny for ablation study of local detail loss and global contour loss. For each setting, we convert the refined sketch and corresponding Canny to binary with a constant threshold value (\emph{i.e.}, each pixel is either zero or one). Precision means how many ones in the refined sketch are actual ones in the ground truth Canny, and recall means how many ones in the ground truth Canny are contained in the refined sketch. The high precision and relatively low recall are in line with expectations. The original purpose of the Stroke Calibration Network is to modify unreasonable strokes and add essential details. The low accuracy of using local detail loss only is consistent with results in Figure \ref{fig:impact_scn}. Since HED and Canny are different, it is not surprising that the accuracy of using global contour loss only is low.

\begin{table}[pos=t]
\centering
\caption{Quantitative performance of our Stroke Calibration Network trained on only local detail loss(detail only), global contour loss(contour only) and local detail loss together with global contour loss(detail and contour).}
\label{tab:impact of SCN}
\begin{tabular}{ccl}
\toprule
Components                   &Precision          & Recall      \\
\midrule
detail only                 & 0.1559            &   0.1475  \\
contour only                & 0.1713            &   0.1564  \\
detail and contour          & \textbf{0.9962}   &\textbf{0.4772} \\
\bottomrule
\end{tabular}
\end{table}

\subsubsection{The scalability of Image Synthesis Network}
\label{sec:scalability of ISN}
Notice that in Section \ref{sec:ISN} we didn't deliberately design the structure of the Image Synthesis Network (SCN). In fact, we adopt a quite simple structure of SCN to show that it is easy to generate satisfactory results from the calibrated sketches even the sktech2image synthesis network is not deliberately designed. Since there are many methods \cite{jo2019sc,yu2018generative} for "well-drawn" sktech2image problem, we argue that such stroke calibration is indispensable for these methods to be well applied in some real applications, such as cultural relics or digital sketch generation for suspects, to produce realistic images. Therefore, it is a useful application and a new solution to synthesize a high-quality image from human-like sketches. The results in Section \ref{sec:comparison against baselines} have shown that our proposed stroke calibration network is a simple yet effective. The calibrated sketches can be directly fed into other existing "well-drawn" sktech2image methods \cite{jo2019sc,yu2018generative} to produce more diverse and more photo-realistic results. Our two-stage Algorithm \ref{alg:Training strategy} provides end-to-end scalability for improving SCN by designing novel architecture or combining with existing "well-drawn" sktech2image methods.

For example, we can improve SCN by simply doubling the numbers of residual blocks (refer to as Improved-1). Or building our generator based on U-Net and using Masked Residual Unit (MRU) module proposed in \cite{Chen2018CVPR}  (refer to as Improved-2). MRU is shown to be more effective than ResNet, Cascaded Refinement Network (CRN) or DCGAN structure in image synthesis task according to \cite{Chen2018CVPR}. We compare images generated by different structures of SCN on the CUHK dataset using PSNR, SSIM and FID as metrics. The results are shown in Table \ref{tab:scalability of ISN}.

\begin{table}[pos=t]
\caption{The scalability of Image Synthesis Network. We compare results generated by different structures of SCN on the CUHK dataset using PSNR, SSIM and FID as metrics. For details refer to Section \ref{sec:comparison against baselines} and Section \ref{sec:scalability of ISN}.}
\label{tab:scalability of ISN}
\begin{tabular}{cccl}
\toprule
Method          &PSNR               & SSIM              & FID\\
\midrule
Original            &{20.25}  &{0.8006} &{58.43} \\
Improved-1         &{20.34}  &{0.8092} &{57.09} \\
Improved-2         &{21.09}  &{0.8137} &{55.12} \\
\bottomrule
\end{tabular}
\end{table}

\section{Conclusion and Discussion}
\label{sec:conclusion}
We propose a human-like sketch based face image synthesis method named \emph{Cali-Sketch}.
Our method can generate pleasing results even when the input sketches are not plausible. To achieve this, we introduce a two-stage sketch-to-image translation method consisting of two GANs. Stroke Calibration Network first calibrates unreasonable strokes and adds necessary details. The refined sketches are then fed into the Image Synthesis Network to produce photo-realistic face images. Given human-like sketches, Cali-Sketch can generate identity-consistent and appearance-realistic face images.
Experimental results show the effectiveness and efficiency of the proposed Cali-Sketch, showing superior performance than the state-of-the-art methods.
\subsection{Analysis for Introducing the Refined Sketch $R$}
In the first place, we try to directly learn the mapping from the sketch to the image using some state-of-the-art methods, \emph{e.g.}, pix2pix~\cite{isola2017image}, CycleGAN~\cite{zhu2017unpaired} and their variants.
However, these existing one-stage methods are unable to generate reasonable and high-quality images directly from the ``badly-drawn sketches''.
The reason is that these one-stage methods are designed for and trained on sketches with precisely and strictly aligned boundaries. 
They necessarily desire a plausible sketch referring to the original image.
When it came to image generation from ``badly-drawn sketches'', which is less discussed but appears more in real applications, these methods might produce unacceptable results given such badly-drawn sketches as input, for example, the restoration results from the open-source demo of SC-FEGAN~\cite{jo2019sc} when the given sketches are badly-drawn, as shown in Figure~\ref{fig:pre-processing_for_SCFEGAN}.
Thus, we divide \textit{image generation from human-like sketches} into a two-stage process: stroke calibration and image generation.
Stroke calibration is solely focused on hallucinating edges in the missing regions. 
The image synthesis network uses the hallucinated edges to generate the final results. 
\par
There are two major benefits by introducing refined sketch $R$.
On the one hand, since there are many methods for the ``well-drawn'' sketch-to-image problem, we argue that such a stroke calibration mechanism is indispensable for the restoration in some real applications like cultural relics or the task of creating digital sketches of suspects, and plays an important role in the final production of realistic images.
In fact, we adopt a simple structure of SCN to show that it is easy to generate a satisfactory result from the calibrated sketches even the network is not deliberately designed.  
The results show that our proposed stroke calibration network is simple yet effective. 
On the other hand, since our stroke calibration adopts the popular HED~\cite{xie15hed} and Canny~\cite{canny1987computational}, the calibrated sketches can be directly fed into other existing ``well-drawn'' sketch-to-image methods~\cite{jo2019sc,yu2018generative,wang2009face,isola2017image} to produce more diverse and more photo-realistic results. 

\subsection{Extended Application}
\begin{figure}[pos=th]
\centering
\includegraphics[width=\linewidth]{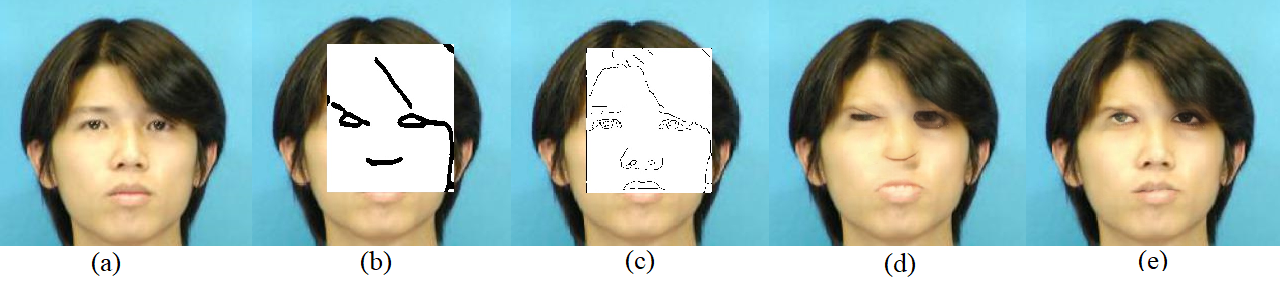}
\caption{Stroke Calibration Network as a pre-processing module of~\cite{jo2019sc} for real-world sketches. (a) original image. (b) masked image and input sketch. (c) masked image and modified sketch. (d) editing result of original sketch. (e) editing result of modified sketch. }
\label{fig:pre-processing_for_SCFEGAN}
\end{figure}

Our Stroke Calibration Network can act as a pre-processing module for real-world sketches. For interactive face image manipulation like \cite{jo2019sc}, a plausible input sketch is necessary. When free-hand drawn sketches are directly fed into those models, the results may be unacceptable. In this case, our Stroke Calibration Network can also act as pre-processing modification module. \cite{jo2019sc} is a recent facial image editing method. Users draw sketch $S$ and color as guidance on incomplete image $I\odot M$ erased by mask $M$. To show the effectiveness and efficiency of our approach, in this case, we first directly use original human-like sketch $S$ as input sketch for \cite{jo2019sc} to get an edited image. Then we feed the refined sketch $R$ pre-processed by our Stroke Calibration Network to produce another edited image. As demonstrated in Figure \ref{fig:pre-processing_for_SCFEGAN}, when the input sketch is sparse and contains wrong strokes and directly fed into \cite{jo2019sc}, the generated facial features are distorted and deformed. Our Stroke Calibration Network can calibrate unreasonable strokes and add necessary details. When this refined sketch is fed into \cite{jo2019sc}, the result is improved significantly.

Our method also show its potential for sketch-based object search \cite{yuan2018query,he2019triangles} and image retrieval \cite{huang2019content,Ioannakis2019Retrieval,Choi2019SketchHelper}. Various works have been proposed to efficiently support automatic annotation of multimedia contents and help content-based retrieval, but obtaining precise image samples sufficing the user specification may not be always handy. In such cases, the sketch can be an alternative solution to initialize the search, \emph{i.e.}, sketch based image retrieval \cite{yuan2018query,Zhang2019retrieval}. Our method can help complete necessary object information critical for a reliable search performance.

\subsection{Limitation and Future Work}
\begin{figure}[pos=th]
  \centering
  \includegraphics[width=\linewidth]{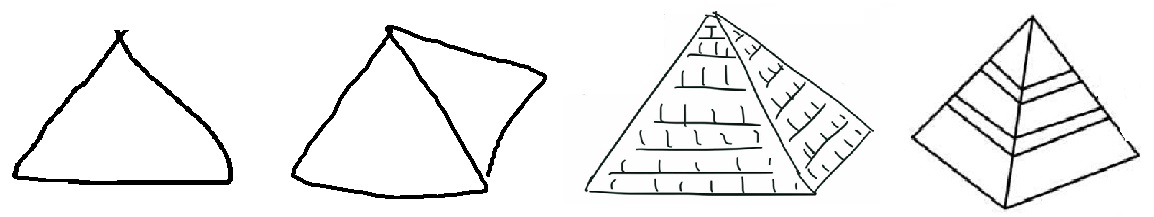}
  \caption{Sketch samples of `triangle'. It is not sufficiently discriminative to uniquely resemble the pyramids by simply drawing a `triangle'. Incorporating category information is critical for image generation from human-like sketches.}
  \label{fig:pyramid}
\end{figure}

Compared with image inpainting \cite{jo2019sc,yu2018generative} or image-to-sketch synthesis \cite{qi2015im2sketch,wang2017data,lu2019fcn}, generating photo-realistic image from human-like sketch is more challenging since there is less information in sketches.
Thus, we temporarily experiment on frontal faces without large rotation and translation. The dataset limitations provide strong motivation for future work to improve performance by expanding the datasets into faces with various angles or expressions, and further into all classes, \emph{e.g.}, the Google QuickDraw Dataset. 
In addition, the category of a sketch is also critical for image generation. Sketches are far from being complete in terms of the object information that would be transformed into a totally different object during generation. For example, as illustrated in Figure \ref{fig:pyramid}, if a user is intent on generating a pyramid image by simply drawing a `triangle', it is not sufficiently discriminative to uniquely resemble the pyramids. 
Thus, incorporating category information or textual descriptions~\cite{xia2020tedigan} of human-like sketches is critical. We will develop our Cali-Sketch into a drawing assistance that creates photographic self-portraits or user's favorite cartoon characters.

\section{Acknowledgements}
This work was partly supported by the Major Research Plan of National Natural Science Foundation of China (Grant No. 61991451), and Shenzhen special fund for the strategic development of emerging industries (Grant No. ZDYBH201900000002).

\bibliographystyle{cas-refs}
\bibliography{cas-refs}

\end{document}